\documentclass[9pt,twocolumn,twoside]{optica2}
\setboolean{shortarticle}{false}
\setboolean{minireview}{false}

\usepackage{import}
\usepackage{blindtext}
\usepackage{cuted}
\usepackage{stfloats}
\usepackage{subeqnarray}

\title{Deep Learning and Model Predictive Control for Self-Tuning Mode-Locked Lasers}

\author[1]{Thomas Baumeister}
\author[2]{Steven L. Brunton}
\author[3,*]{J. Nathan Kutz}

\affil[1]{Technical University of Munich, Arcisstra${\beta}$e 21,  D-80333 Munich, Germany}
\affil[2]{Department of Mechanical Engineering, University of Washington, Seattle, WA 90195  USA}
\affil[3]{Department of Applied Mathematics, University of Washington, Seattle, WA 90195-3925  USA}

\affil[*]{Corresponding author: kutz@uw.edu}

\ociscodes{(140.4050) Mode-locked lasers; (140.3510) Lasers, fiber; (150.5758) Robotic and machine control; (120.4820) Optical systems}

\begin{abstract}
Self-tuning optical systems are of growing importance in technological applications such as mode-locked fiber lasers.  Such self-tuning paradigms require {\em intelligent} algorithms capable of inferring approximate models of the underlying physics and discovering appropriate control laws in order to maintain robust performance for a given objective. 
In this work, we demonstrate the first integration of a {\em deep learning} (DL) architecture with {\em model predictive control} (MPC) in order to self-tune a mode-locked fiber laser.  Not only can our DL-MPC algorithmic architecture approximate the unknown fiber birefringence, it also builds a dynamical model of the laser and appropriate control law for maintaining robust, high-energy pulses despite a stochastically drifting birefringence.   We demonstrate the effectiveness of this method on a fiber laser which is mode-locked by nonlinear polarization rotation.  The method advocated can be broadly applied to a variety of optical systems that require robust controllers.
\end{abstract}

\setboolean{displaycopyright}{true}

\begin{document}

\maketitle

\section{Introduction}

Data-driven modeling of physical systems is leading to new paradigms for the engineering of intelligent, self-tuning systems.  Enabled broadly by machine learning and artificial intelligence algorithms, engineering design principles and control laws can be extracted from data alone, including latent variables that can be inferred and predicted without direct measurements~\cite{Fu2014oe}.  
%
%
Many nonlinear optical systems are ideally suited for performance enhancements through such machine learning methods.  For instance, {\em mode-locked fiber lasers} (MLFLs), which will be an exemplar of the ideas advocated in the following, are continuing to achieve significant performance gains through innovative engineering design~\cite{richardson2010high}.  However, the performance gains produce MLFLs that are not robust to environmental and parametric laser cavity disturbances.  Control is also difficult to achieve due to the strong cavity nonlinearity, its nearly instantaneous response to disturbances, and unknown cavity parameters such as fiber birefringence.  In this manuscript, we show that the flexible and adaptive control architecture of {\em model predictive control} (MPC) can be coupled with the industry leading machine learning method of deep neural networks ({\em deep learning} (DL)) in order to produce a robust self-tuning system.  Our data-driven paradigm thus integrates state-of-the-art machine learning and control algorithms to produce a highly advantageous, and easy to implement, learning module for self-tuning of optical systems.

Optical systems are a ubiquitous technology, especially laser systems which are the backbone of an \$11B/year and growing industry~\cite{lasermarket}.  In particular, there is a growing demand for emerging high-power MLFL technologies, which have the potential to close the performance gap with expensive solid state lasers while using readily available telecommunications components.  Compact and inexpensive, MLFLs are a turn-key technology capable of revolutionizing commercial applications and scientific efforts.  The dominant commercially available MLFLs are based upon {\em nonlinear polarization rotation} (NPR) that occurs in a birefringent fiber cavity with waveplates and a polarizer~\cite{hausreview,kutzreview}.  Due to strong nonlinear effects, the performance of the NPR based laser is highly sensitive and is subject to significant performance losses with only small changes to the cavity or environment, thus requiring stringent engineering tolerances and environmental shielding to achieve optimal performance.  To overcome such cavity sensitivity, adaptive control~\cite{Krstic:2000} and machine learning algorithms for self-tuning have been proposed~\cite{Brunton2013jqe,Fu2014oe,Brunton2014ieeejstqe} and enacted experimentally~\cite{Andral2015optica,andral2016toward,Woodward2016sr} using servo-driven waveplates and/or polarizers~\cite{shen2012electronic,radnatarov2013automatic}.  The algorithms include methods for extracting the birefringence, which is a latent variable and cannot be measured directly~\cite{Fu2014oe}.  These machine learning algorithms demonstrate the promise of data-driven methods for engineering robust and self-tuning MLFLs.  
More generally, machine learning and adaptive control have recently been applied to a host of optical systems, including optical communication~\cite{zibar2016machine,khan2016modulation,wang2017failure}, laser cutting~\cite{zibar2016machineb,tercan2017improving}, metamaterial antenna arrays~~\cite{Johnson2015josaa}, microscopy~\cite{albert2000smart}, and characterization of x-ray pulses~\cite{sanchez2017accurate}.  

Although machine learning has already been applied to improve robust performance in optical systems, emerging  algorithms continue to push the limits of what can be achieved.  Foremost among emerging data methods are deep neural networks (DNNs).   DNNs have dominated the machine learning landscape in data-rich applications, as they have been demonstrated to achieve superior performance in comparison to other techniques by almost any meaningful metric.  They are also capable of inferring latent variables and adaptive learning, both features that are exploited in our algorithm design.   DNNs have also generated a diversity of readily available software (See, for instance, Google's open source software called TensorFlow:  tensorflow.org).   

Like DNNs, MPC has been used in process control since the 1980s.  In recent years, it has emerged as a leading control architecture due to its flexible and robust performance on strongly nonlinear systems~\cite{garcia1989model}.  As such, it has started to dominate much of the engineering and commercial market, showing great success in handling complex constrained control problems \cite{XI.2013}.  
The success of MPC is based on the fact that the current control action is optimized for performance over a finite time-horizon subject to system constraints.  
As the system advances in time, the optimization is repeatedly performed for each new timestep, and the control law is updated.  
MPC performance gains are thus a direct consequence of the predictive capabilities and control updates which come at the cost of an optimization step.  As such, MPC is widely implemented in research and industry.  

In this work, we integrate the state-of-the-art, and highly advantageous, methods of DL and MPC to develop a  {\em Deep learning model predictive control} (DL-MPC) algorithm for self-tuning, intelligent optical systems.  We demonstrate the DL-MPC architecture on MLFLs, showing that our method can learn a model for the unknown birefringence, and can then use this information to keep the MLFL at peak performance despite parametric laser cavity disturbances.  The algorithm is flexible and capable of building robust, physics-based models for control.  The DL-MPC for MLFL provides a compelling architecture for modern intelligent systems, especially systems like MLFL where strong nonlinearities compromise standard controllers.


\section{Background:  Deep Learning and Model Predictive Control}
\label{sec:examples}

The work detailed here integrates two important theoretical concepts.  We provide a brief review of each method in order to highlight how they are connected in our laser control model.

\subsection{Development of Model Predictive Control}

MPC provides a robust architecture for explicitly incorporating constraints in control laws.  This is of growing importance in many industrial applications where rapid technological progress has created increased demand for control laws that can easily handle the constraints imposed by the integration of a variety of technological components.  The receding horizon strategy of MPC enables the incorporation of the constraints by enforcing the constraints on future inputs, outputs and state variables \cite{XI.2013}.   MPC makes use of an explicit dynamic plant model to predict future outputs of the system $\hat{\boldsymbol{y}}_{t+1:t+N-1}$ for the prediction horizon $N$ given the future control inputs $\boldsymbol{u}_{t:t+N-1}$. The control inputs are optimized so that the system's cost function $J$ is minimized. The cost function must incorporate the objective of the control law. The optimization of $J$ subject to the plant model and the constraints can be solved using (nonlinear) quadratic programming.   The success of MPC is a direct consequence of the optimization procedure, allowing it to supplant standard LQR and PID controllers in many situations. 

MPC and its enhancements have been successfully implemented in a large variety of applications~\cite{garcia1989model,lee2011springer}, such as control of a Gasoline HCCI Engine, flight control, and satellite attitude control. 
Despite the great success and the rapid development of MPC, there are still challenges limiting the applicability to many industrial systems \cite{XI.2013,Mohanty.2009}. While the ability to incorporate constraints in the control optimization is a primary advantage of MPC, it is computationally expensive and requires high performance computing to execute.  Additionally, the performance of the controller strongly depends on the ability of the plant model to capture the dynamics of the system \cite{Mohanty.2009}. This makes the plant model development the most critical and time-consuming part of designing an MPC architecture \cite{WeisbergAndersen.1992}.  However, machine learning methods are able to extract engineering dynamics from spatio-temporal data alone and, thus, are well-suited to overcome these limitations.

\subsection{Deep Neural Networks and Learning}

Deep learning provides a powerful mathematical framework for supervised learning \cite{Goodfellow-et-al-2016}.  It can also be successfully modified for unsupervised model building with \emph{reinforcement learning} (RL) and \emph{variational autoencoders} (VAEs) \cite{Mnih.2015,Danilo.2014}.  RL, VAEs and convolutional neural networks (CNNs) form the key algorithmic structures of deep learning.  One of its most compelling technological milestone was reached when a recent deep learning algorithm was able to defeat a human professional player in the game of Go \cite{Silver.2016}. This achievement was previously thought to be at least a decade away.  It further illustrates the ability of DNNs to successfully learn and execute nuanced tasks.

The recent success of DNNs has been enabled by two critical components:  (i) the continued growth of computational power, and (ii) exceptionally large labeled data sets which take advantage of the power of a multi-layer (deep) architecture. It was the analysis and training of a DNN using the ImageNet data set in 2012~\cite{Krizhevsky2012nips} that provided a watershed moment for {\em deep learning}~\cite{Lecun2015nature}.  DNNs have since transformed many fields by dominating the performance metrics in almost every meaningful task intended for classification and identification.
Interestingly, DNNs were not even listed as one of the top 10 algorithms of data mining in 2008~\cite{top10}.  But in 2017, its undeniable and growing list of successes make it perhaps the most important data mining tool for our emerging generation of scientists and engineers.

DNNs are rooted in foundational and rigorous mathematics.   Indeed the proof of the universal approximation theorem \cite{Cybenko.1989,Hornik.1989,Hornik.1990} showed that a DNN with a linear output layer and a sufficient number of hidden units was able to learn any arbitrary function.  However, there is no quantification of how many hidden units are required or guarantee that the training algorithm can find the correct, globally optimal, model parameters \cite{Goodfellow-et-al-2016}.  Training algorithms can also often fail due to overfitting or when computation of gradients vanish or blowup \cite{Goodfellow-et-al-2016}.  Overfitting renders the model valid only for the specific data trained on and does not provide generality for the model, but can be avoided by cross-validation and early stopping of the optimization routine \cite{Murphy.2012}.   The vanishing gradient problem occurs while backpropagating the error from the output layer towards the input layer. The further away from the output layer, the smaller the gradients become and, consequently, the smaller the updates of the model parameter are. An efficient way to reduce the vanishing gradient problem is to use rectified linear units (RLU) $a(z) = max\{0,z\}$ as activation functions. RLU units retain a large gradient if the unit is active. In contrast, the exploding gradient problem occurs in a DNN when the forward propagation of the feature input results in large differences between the predicted output and the true output. These large differences, in turn, lead to large gradients of the output layer. Backpropagating such large gradients can cause instabilities in the iterative structure of the optimization while learning \cite{Hanie.2016}. One approach to mitigate this effect is to implement pre-learning steps to ensure good initial model parameters.  Despite the difficulties in the  initialization and optimization procedure, DNNs have been successfully trained in a large variety of applications areas with impressive performance.

\subsection{Integration of Deep Learning in Model Predictive Control}

The success of MPC and DNNs for learning can be integrated to provide a compelling and robust control architecture.
Specifically, MPC is ideally constructed for a constrained optimization control architecture.  
However, MPC requires an accurate dynamical plant model.  By using the DL to learn an accurate plant model, the DL and MPC architectures can be naturally integrated.  This leverages the advantage of both algorithms and provides a robust mathematical architecture for control.  To date, there have been a few research teams which successfully integrated machine learning methods such as radial basis functions (RBFs), gaussian processes (GPs) and DNNs with the MPC architecture \cite{Peng.2009,Grancharova.2008,Tsai.2009}. The DL-MPC  provides a new viewpoint for controlling optical systems with machine learning.

\section{Deep Learning for MLFL Control}
\label{sec:Deep Learning ML}

\begin{figure}[t]
\centering
\includegraphics[width=.9\linewidth]{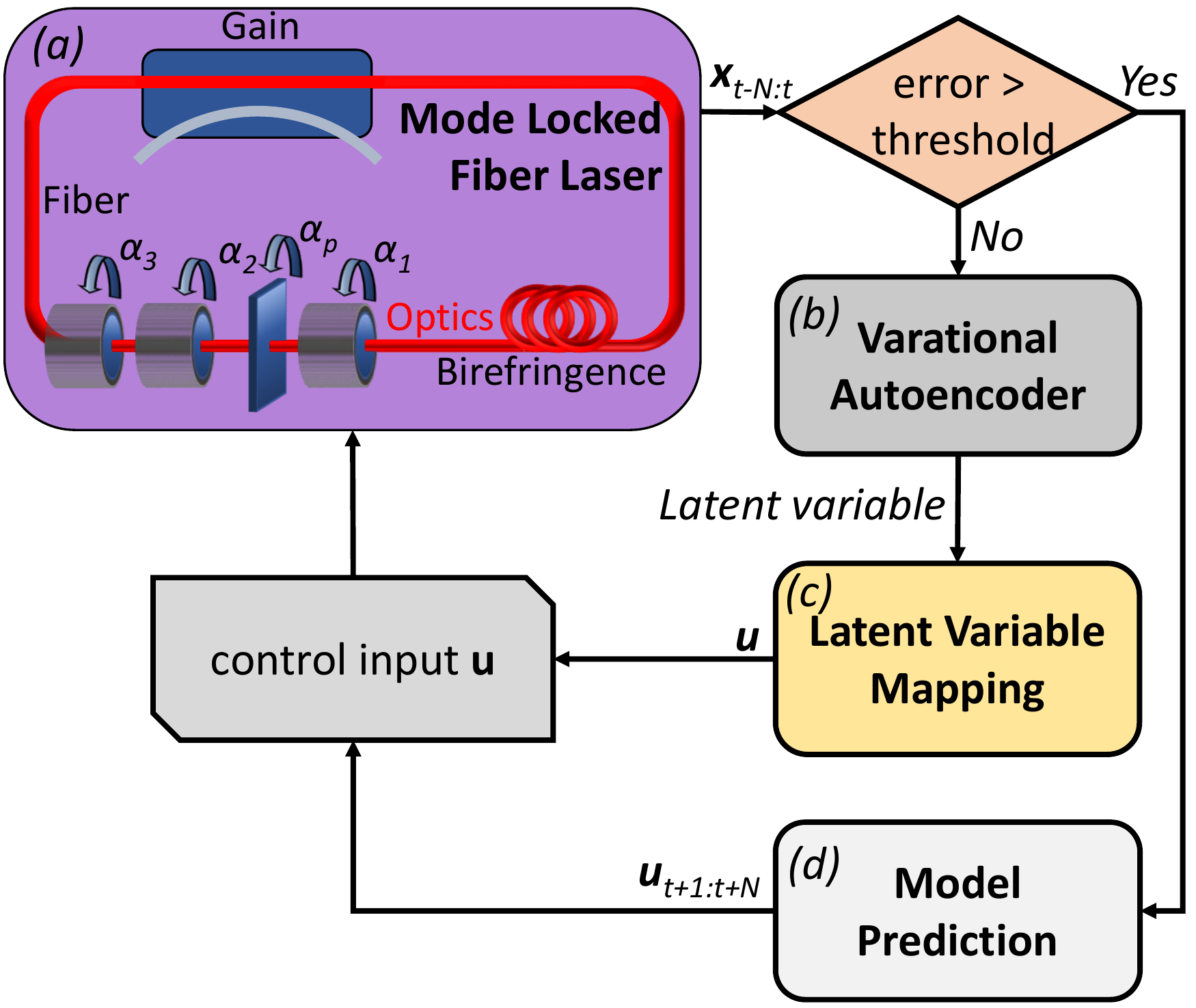}
\vspace{-.1in}
\caption{Schematic of the self-tuning fiber laser. The laser cavity, optic components and the laser's objective
function $(a)$ are discussed in section \ref{sec:Mode-locked Fiber Laser}. The Variational Autoencoder $(b)$ is 
discussed in section \ref{sec:Variational Autoencoder}, the Latent Variable Mapping $(c)$ in section 
\ref{sec:Latent Variable Mapping} and the Model Prediction $(d)$ in section \ref{sec:Model Prediction}.}
\vspace{-.15in}
\label{fig:Control_Model}
\end{figure}

A schematic of the self-tuning MLFL is shown in Fig. \ref{fig:Control_Model}. The ML-MPC consists of a $(b)$ VAE for inferring the birefringence, $(c)$ a latent variable mapping, and $(d)$ a model prediction. The MLFL device itself $(a)$, which is controlled by waveplates and polarizers, as well as each component of the ML-MPC are discussed in the following subsections.   Our key innovations of $(b)-(d)$ are based upon integrating a number of statistical methods which sample the laser behavior and infer both a model for the birefringence and the cavity dynamics.

For any such data-driven strategy to be effective, we require an objective function $O$, with local maxima that correspond to high-energy mode-locked solutions.  Although we seek high-energy solutions, there are many chaotic waveforms that have significantly higher energy than mode-locked solutions.  Therefore, energy alone is not a good objective function.  Instead, we divide the energy function $E$ by the fourth-moment (kurtosis) $M$ of the Fourier spectrum of the waveform
\[
O = {E}/{M}
\]
which is large for undesirable chaotic solutions.  This objective function, which has been shown to be successful for applying adaptive control, is large when we have a large amount of energy in a tightly confined temporal wave packet~\cite{Brunton2013jqe,Fu2014oe,Brunton2014ieeejstqe}.

\subsection{Mode-locked Fiber Laser Model and Data}
\label{sec:Mode-locked Fiber Laser}

This section outlines the model used for generating data that characterizes the MLFL.  In practice, the data acquisition would be directly from experiment.
The intra-cavity dynamics of the mode-locked laser must account for, among other things, 
the nonlinear polarization dynamics and energy equilibration responsible for initiating the mode-locking process.
Although we consider the passive polarizer and waveplates as discrete elements in
the laser cavity, the remaining physical terms are lumped together into an averaged propagation equation
that includes the chromatic dispersion, Kerr nonlinearity, attenuation, and bandwidth-limited, saturating gain~\cite{edwin1,edwin2}: 
\begin{subeqnarray} 
&& \hspace{-.4in} i \frac{\partial u}{\partial z} \!+\! \frac{D}{2} \frac{\partial^{2}u}{\partial t^{2}}\!-\! Ku \!+\! (|u|^{2} \!+\! A|v|^{2})u \!\!+\! Bv^{2}u^{*} \\
&&\hspace*{1in} \!\!=\! i g(z)\left(1+\tau \frac{\partial ^{2}}{\partial t^{2}} \right) u - i \Gamma u  \nonumber \\
&& \hspace{-.4in} i \frac{\partial v}{\partial z} \!+\! \frac{D}{2} \frac{\partial^{2}v}{\partial t^{2}} \!+\! Kv \!+\!(A|u|^{2} \!+\! |v|^{2})v \!\!+\! Bu^{2}v^{*} \\
&& \hspace*{1in} \!\!=\! i g(z)\left(1+\tau \frac{\partial ^{2}}{\partial t^{2}} \right) v - i \Gamma v  .\nonumber
\label{eq:cnls}
\end{subeqnarray}
The left hand side of this equation is the coupled nonlinear Schr\"odinger equations (CNLS).
This system models the averaged propagation of two orthogonally polarized electric field envelopes in a birefringent optical fiber in  nondimensionalized form for which the $u$ and $v$ fields 
are orthogonally polarized components of the electric field.   
The right hand side terms arise from saturated, bandwidth-limited gain given by
\begin{equation} 
g(z) = \frac {2g_{0}}{1 + \frac{1}{E_0} \int_{-\infty}^{\infty} (|u |^{2} + |v |^{2}) \mbox{d}t} \, ,
\label{eq:g} 
\end{equation}
and linear attenuation (cavity losses). 
Here $g_{0}$ and $E_0$ represent the gain (pumping) strength and cavity saturation energy respectively, while
$\Gamma$ models the distributed losses due to output coupling, fiber attenuation, splicing and interconnects.   

The variable $t$  represents the physical time in the rest frame of the pulse normalized by $T_0$, where 
$T_0$ (e.g. 200-300 femtoseconds) is the full width at half-maximum of the mode-locked pulse and the 
variable $z$ is the physical distance divided by the length $L$ of the laser cavity.
We have scaled the complex orthogonal fields $u$ and $v$ by the 
factor $\sqrt{\gamma L}$ where $\gamma=n_2 \omega_0 / (c A_{\mbox{eff}})$.
Here $n_2=2.6\times 10^{-16}$ cm$^2$/W is the nonlinear index coefficient of 
the fiber, $\omega_0 \approx 10^{15}$ s$^{-1}$ is the center frequency of the pulse spectrum for a pulse at $\lambda_0=1.55 \mu$m, 
$c$ is the free-space speed of light, and $A_{\mbox{eff}}=55\mu$m$^2$ is the average cross-sectional area of the fiber cavity.
The parameter $D$ characterizes the averaged normal ($D<0$) or anomalous ($D>0$) chromatic dispersion in the laser cavity.
Specifically, in the normalizations used here $D =\beta_{2} L / L_{D}$ where $\beta_{2}$ (in ps$^2$ m$^{-1}$) is the averaged 
dispersion of the fiber, and $L_{D}$ is the dispersion length defined by $T_{0}^{2} / \left| \beta_{2} \right|$.
The birefringence strength parameter $K$ determines the effective relative phase velocity difference between the 
$u$ and $v$ fields. The material properties of the optical 
fiber determine the values of nonlinear coupling parameters 
$A$ and $B$ which satisfy $A+B=1$ by 
axisymmetry, specifically $A=2/3$ and $B=1/3$. 

With establishment of the intra-cavity propagation dynamics, it only remains to apply the discrete effects of the
waveplates and passive polarizer in the laser cavity to induce mode-locking.
Jones matrices are used to model the effects of waveplates and polarizer~\cite{jones}. 
When the principle axes of these devices are aligned with the fast axis of the fiber, the Jones matrices of 
the quarter-waveplate, half-waveplate and polarizer are
\begin{equation}
\hspace*{-0.1in}
 W_{\frac{\lambda}{4}} \!=\!\! \left(\!\!\! \begin{array}{cc} e^{-i \pi / 4} & 0 \\ 0 & e^{i \pi / 4} \end{array}\!\!\! \right)\!, 
 W_{\frac{\lambda}{2}} \!=\!\! \left(\!\! \begin{array}{cc} -i & 0 \\ 0 & i \end{array}\!\! \right) \!,  
  W_{p} \!=\!\! \left(\!\! \begin{array}{cc} 1& 0 \\ 0 & 0 \end{array}\!\! \right) \!.
\label{eq:jones1}
\end{equation}
%
For arbitrary orientation $\alpha_j$ (see Fig.~\ref{fig:Control_Model}(a)) with respect to the fast axis of the fiber, the above matrices are modified by:
\begin{equation}
J_{j} =  R(\alpha_{j}) W R(-\alpha_{j})
\label{eq:jones}
\end{equation}
where $W$ is one of the matrices in Eq.~(\ref{eq:jones1}) and $R(\alpha_j)$ is a standard rotation matrix of angle $\alpha_j$.
%
To help make clear the model of the laser cavity dynamics subject to Eq.~(\ref{eq:cnls}), consider a
single round trip in the cavity.
The propagation of the field starts right after the polarizer with orientation $\alpha_{p}$ for which the pulse is linearly polarized. The quarter-waveplate (with angle $\alpha_{1}$) to the left of the polarizer converts the polarization state from linear to elliptical, thus creating a polarization ellipse. Upon passing through the laser cavity, the polarization ellipse is subjected to a intensity-dependent rotation as well as
amplification as governed by~(\ref{eq:cnls}).  At the end of fiber, the half-waveplate (with angle $\alpha_{3}$) further rotates the polarization ellipse. The quarter-waveplate (with angle $\alpha_{2}$) converts the polarization state from elliptical back to linear, and the polarizer finally aligns the field with its own principle axis.

\subsection{Variational Autoencoder}
\label{sec:Variational Autoencoder}

Mode-locking is highly sensitive to the birefringence of the MLFL which cannot be directly  measured, i.e. it is a latent variable.  Moreover, every fiber draw produces a unique, stochastic realization of the birefringence.  Compounding this is the fact that moving the fiber, and/or temperature fluctuations, also changes the birefringence.  The physical effects of birefringence have significantly limited any quantitative characterization of MLFLs~\cite{Fu2014oe}.  
A VAE is used to infer a representation of the birefringence from its latent space since recent work has shown that a VAE is able to learn a meaningful structured latent space \cite{Kingma.2014b,Kulkarni.2015}. 

The VAE is a generative model rooted in Bayesian inference, i.e. it estimates the underlying probability distribution so that new data $x$ can be sampled from that distribution:
\begin{equation}
p(x) = \int\limits_{-\infty}^{\infty} p(x|z)p(z)dz = \frac{p(x|z;\theta)p(z)}{p(z|x;\theta)},
\end{equation}
where $z$ are samples from the stochastic latent space $\mathcal{Z}$. Unfortunately, computing this integral numerically takes exponential time to be evaluated over all configurations. Instead, Bayes' theorem is applied to rewrite this integral, where $p(z|x;\theta)$ is the posterior distribution. This distribution can be approximated by $q(z|x;\phi)$. The Kullback Leibler (KL) divergence measures how much information is lost when using the approximation:
\begin{equation}
KL(q(z|x;\phi)|| p(z|x;\theta)) = ELBO(\phi, \theta) + \log p(x), \label{eqn:KL}
\end{equation}
where the evidence lower bound (ELBO) is defined as:
\begin{equation}
ELBO(\phi, \theta) = \mathbb{E}_q[\log p(z|x;\theta)] - \mathbb{E}_q[\log q(z|x;\phi)].\label{eqn:ELBO}
\end{equation}

The objective is to minimize the KL divergence such that the approximation is optimized:
\begin{equation}
q^*(z|x;\phi) = arg min KL(q(z|x;\phi)||p(z|x;\theta)).
\end{equation}
%
This cannot be solved directly since the evidence $p(x)$ is part of the KL divergence. However, it is proven that $KL \geq 0$ using Jensen's inequality \cite{Cover.1991}. Making use of this property and that the logarithm is a monotonic function, it can be shown that minimizing the KL divergence is equivalent to maximizing the ELBO. The ELBO can be rewritten and decomposed to be dependent on a single data point:
\begin{equation}
ELBO_i(\phi,\theta) = \mathbb{E}_q[\log p(x_i|z;\theta)] - KL(q(z|x_i;\phi)|| p(z)).
\end{equation}
Since the loss function in a NN will be always minimized, $\mathcal{L}$ is defined as the negative ELBO. To be able to backpropagate this loss through the VAE with a stochastic latent space, a reparameterization has to be applied. The reparametrization defines $z = \mu + \sigma\odot\varepsilon$, where $\varepsilon$ is a sample from $\mathcal{N}(0, 1)$ and $\odot$ signifies an element-wise Hadamard product. Thus, the randomness of the latent variable is shifted into $\varepsilon$. 
%
This loss function has been shown to ensure a meaningful structure of the latent space \cite{Kingma.2014,Danilo.2014} when estimating the underlying probability distribution. The implemented VAE is a modification of \cite{Kingma.2014}.

\subsection{Latent Variable Mapping}
\label{sec:Latent Variable Mapping}

The latent variable mapping is a simple fully connected NN that maps the representation of the non-measurable parameters $K$ to a good initial control input $\boldsymbol{u}$. Depending on the complexity of the mapping, the network architecture can be adjusted, i.e. increasing or decreasing the number of hidden layers and the number of neurons in each layer. The loss function is defined as the $L_2$ norm, which is commonly for least-squares regression:
\begin{equation}
\mathcal{L} = \frac{1}{2} ||\hat{\boldsymbol{u}} - \boldsymbol{u}||^2_2.
\end{equation}
%
The stochastic optimization algorithm Adam \cite{Kingma.2015} was used to train the model.

The critical part of the latent variable mapping is to ensure that the birefringence $K$ is mapped to good control inputs $\boldsymbol{u}$ which maintain the objective function at a high level. To do so, the DNN is trained on a subset of simulation data of Eq.~(\ref{eq:cnls}). This subset is identified in the following way. First, the interval $[K_{min}, K_{max}]$ is divided into $n$ equidistant parts $\boldsymbol{K}_{subset}$. Within the range $\pm \delta K$ around $K_{subset,i}$ the control inputs $\boldsymbol{u}^*_i$ from the data set with the highest corresponding objective function are appended to the subset, for $i = 1, \dots, n$.

\subsection{Model Prediction}
\label{sec:Model Prediction}

The model prediction (MP) module is the centerpiece of the DL-MPC architecture. This recurrent neural network (RNN) undertakes the task of a classic MPC by first predicting the birefringence and the laser states $N$ time steps in the future and, second, optimizing the future control inputs such that the objective function $O=E/M$ is maximized:
\begin{equation}
\underset{\boldsymbol{u}_{t+1:t+N}}{arg\,max \text{ }} O_{t+1:t+N} = \underset{\boldsymbol{u}_{t+1:t+N}}{arg\,max \text{ }} \left\{\frac{E}{M}\right\}_{t+1:t+N}.
\end{equation}
The MP consists of an encoder and a decoder with $N$ cells, respectively. While the encoder gathers information about long term dynamics, the decoder performs the actual prediction task.

\begin{figure*}[t]
	\centering
	\includegraphics[width=0.8\textwidth]{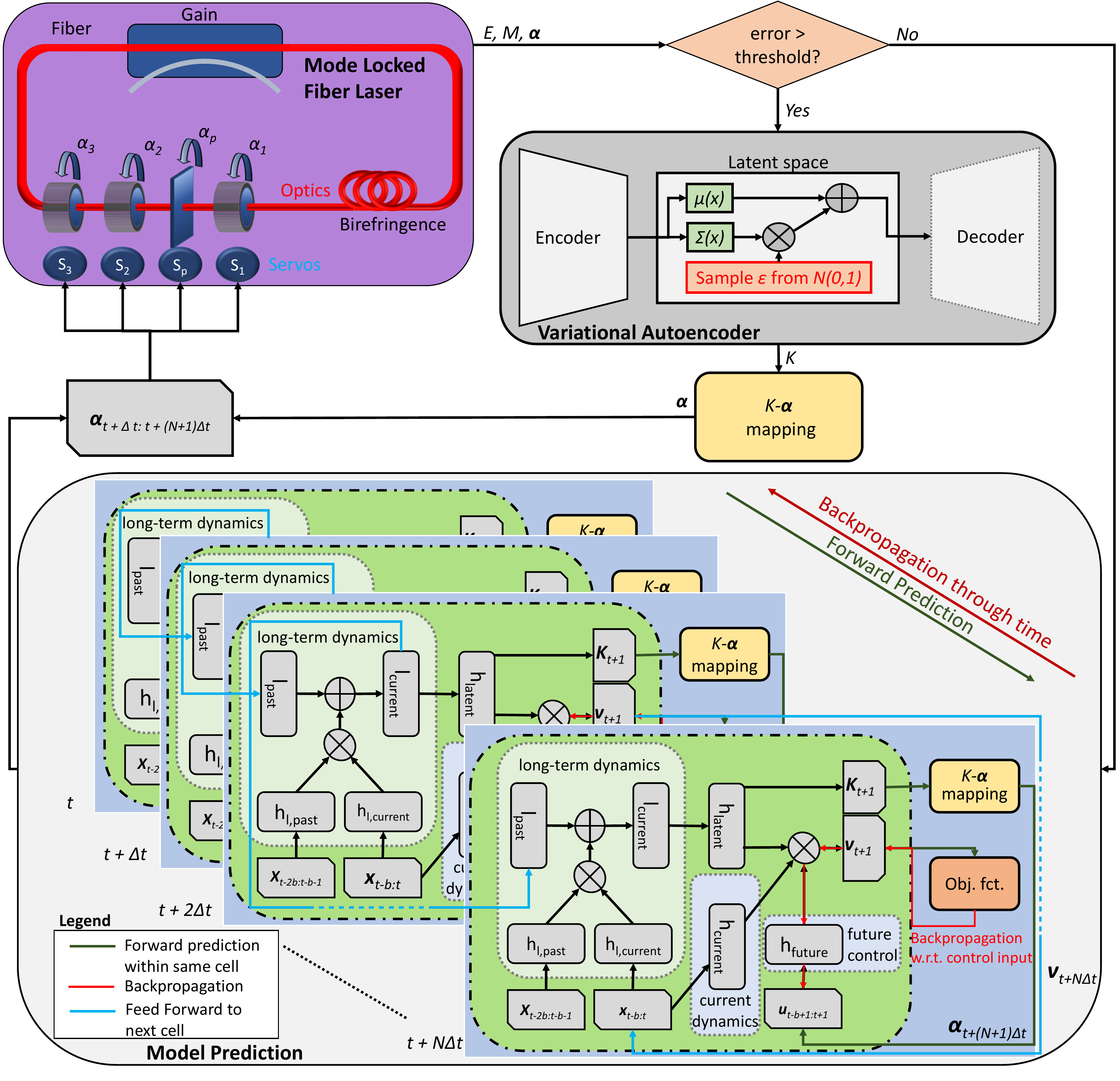}
	\vspace*{-.15in}
	\caption{Schematic of the Deep Learning Controller. The inputs to the controller 
     are sequences of the states of the laser $E$, $M$ and of the control inputs $\boldsymbol{\alpha}$.
     The Model Prediction is a RNN that first predicts the birefringence $K_{t+1}$ of time step $t+1$ and 
     maps it to good initial control inputs $\boldsymbol{\alpha}_{t+1}$. Second, the system's states $v_{t+1}$ are predicted.
     This is done recurrently to predict $N$ time steps in the future. Then, the control inputs are updated
     such that the objective function is optimized. The optimized control inputs $\boldsymbol{\alpha}_{t+\Delta t: t+ N\Delta t}$
     are used to regulate the laser system for the next $N$ time steps. Once the difference between the prediction 
     and the true output exceed a certain threshold, the VAE is used to infer $K$ and, then, \texttt{K-$\boldsymbol{\alpha}$ mapping} 
     maps it to the control input $\boldsymbol{\alpha}$. This inner loop is necessary to stabilize the control system.}
     \label{fig:Laser_System}
\end{figure*}

\paragraph{RNN Cell}

Each cell $k$ takes as input the sequences $\boldsymbol{x}_{t-2b+k:t+k} = \{E,M,\boldsymbol{\alpha}\}_{t-2b+k:t+k}$ and a sequence of control inputs,
for $k = 1, \dots , 2N$. Here, $b$ is a hyperparameter defining the sequence length of the inputs. The cells are divided into three functional parts capturing different types of the system's dynamics, i.e. long term dynamics, current dynamics and the influence of the control inputs (see Fig.~\ref{fig:Laser_System}).  Details about the equations for the RNN cells are provided in the Appendix.  


\paragraph{Training of the Model Prediction RNN}

The training of RNNs has been the subject of considerable effort in the past decades~\cite{Williams.1995,Hanie.2016}. There are two key challenges:  (i) the exploding and the vanishing gradient effects, which are especially critical for problems with long-range dependencies, and (ii) the effect of nonlinearities when iterating from one time step to another~\cite{Hochreiter.1997,Martens.2011}.
%
It has been shown that good initial model parameters have a tremendous effect on overcoming these difficulties and the ultimate success of training an RNN.  This work implements a three-stage pre-learning approach, similar to that developed in \cite{Lenz.2015}:

\begin{enumerate}
\item \textit{Conditional Restricted Boltzmann Machines} (CRBMs) play an important role for training RNNs by computing good initial model parameters \cite{Hinton.2006,Hinton.2006b}. A classic RBM is an energy-based model that has a layer of visible units $\boldsymbol{v}$ and a layer of hidden units $\boldsymbol{h}$. The goal of the RBM is to learn a set of model parameters $\boldsymbol{\Theta}$ such that the reconstructed units $\boldsymbol{v}_{recon}$, which are propagated to the hidden layer and, then, back to the visible layer, preserving as much information about $\boldsymbol{v}$ as possible. Typically, RBMs use logistic units for both the visible and hidden units. However, in our case we assume a continuous dynamical system and, thus, a modified RBM is implemented in which the $\boldsymbol{v}$ are linear real-valued variables that have Gaussian noise \cite{Freund.1994,Welling.2005}. Incorporating temporal dependencies makes this model a conditional RBM. This pre-learning step is used to compute the model parameters connected to an input layer.

\item \textit{Single-Step Prediction} reduces the range of temporal dependencies and, consequently, the effect of exploding gradients which facilitates the training. For the remaining weights, which were not computed by CRBM, Xavier initialization was used. This method keeps the signal propagating through the network in a reasonable range of values even for many layers \cite{Xavier.2010}.  The trained model parameters are a good basis for the receding horizon prediction.

\item \textit{Receding Horizon Prediction} feeds the output $\boldsymbol{v}$ of a decoder cell forward as the input of the next one. By applying this recurrently, several time steps in the future can be predicted. In this work, the receding horizon was chosen to be $N = 10$. To optimize this system a variation of the backpropagation through time (BPTT) algorithm was used and the loss function was defined as the sum-squared prediction error up to $N$ time steps in the future.
\end{enumerate}

\begin{figure}[t]
\centering
\includegraphics[width=\linewidth]{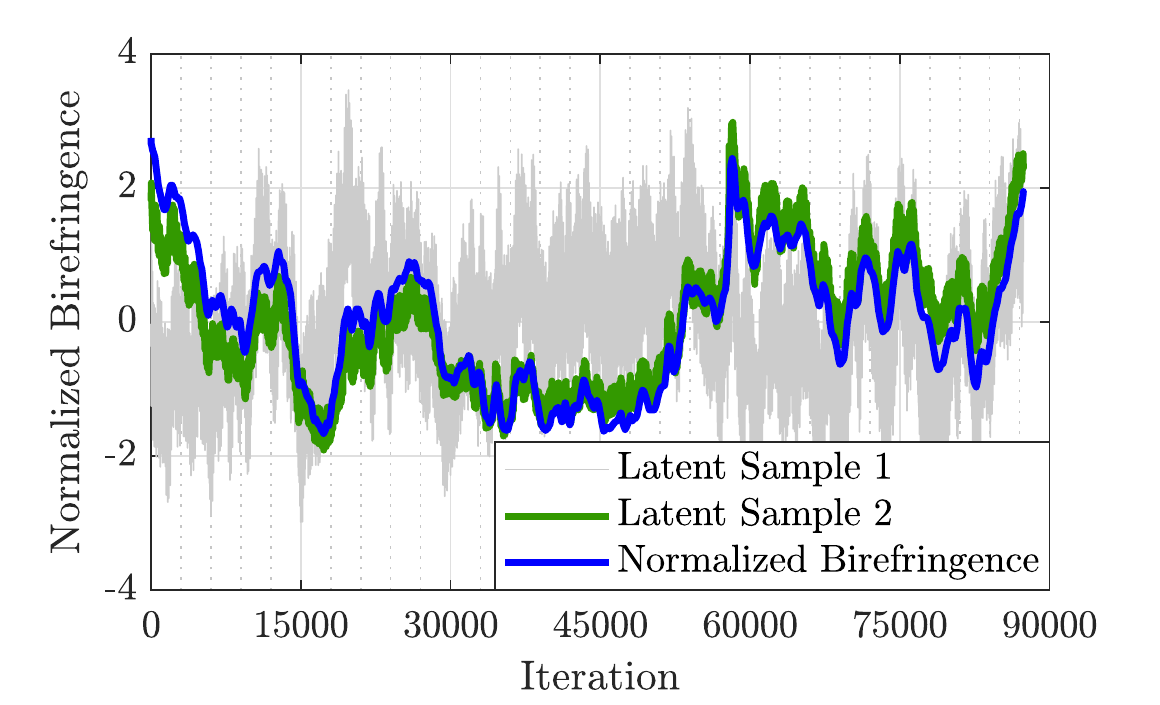}
\vspace{-.25in}
\caption{Comparison of the true birefringence (blue line) and the samples from the two dimensional 
VAE's latent space. While the samples from the first dimension seem to capture just random noise, 
the samples from the second dimension follow the true birefringence with high accuracy.}
\label{fig:VAE_birefringence}
\vspace{-.2in}
\end{figure}
\begin{figure}[t]
\vspace{-.1in}
\centering
\includegraphics[width=\linewidth]{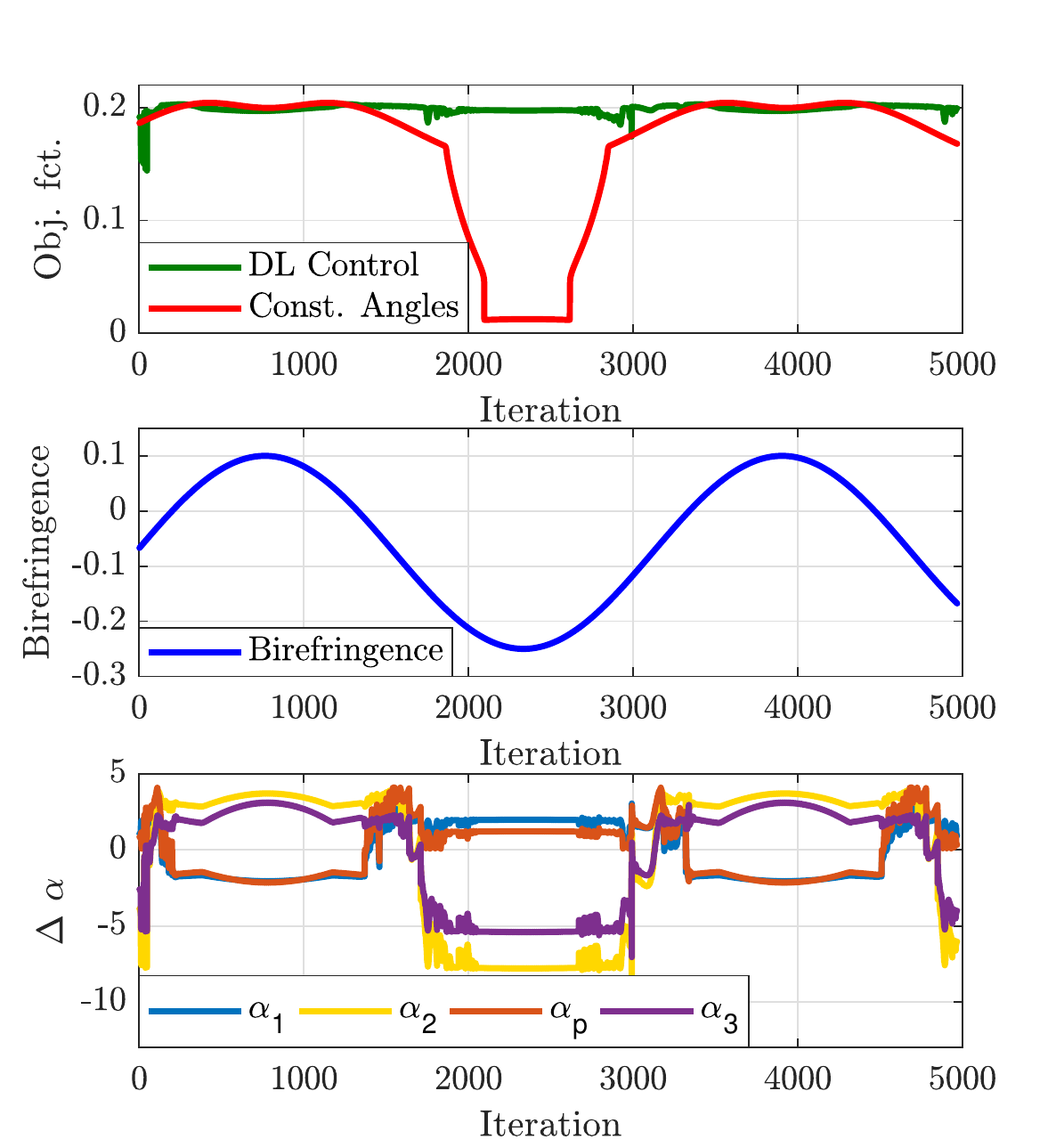}
\vspace{-.25in}
\caption{Performance of the Deep Learning Control despite significant sinusoidal change in birefringence
over time. Without control, the objective function plummets and results in failure of the fiber laser to mode-lock. 
With DL-MPC, the system remains at a high-performance mode-locked state.}
\label{fig:sinusoidal}
\end{figure}

\paragraph{Deep Learning Control Algorithm}

The DL-MPC consists of an inner and an outer loop (see Fig. \ref{fig:Laser_System}). The inner loop includes the VAE and the $K$-$\boldsymbol{\alpha}$ mapping and the outer loop the MP-RNN. Since the MP-RNN needs at least temporal information from $N+2b$ time steps ($N$ encoder cells and input sequence of $2b$), the inner loop has to control the MLFL at the beginning.  
The measured laser states and the control inputs $(E_t, M_t, \boldsymbol{\alpha}_t)$ are fed in the VAE to sample $K_t$, which in turn is mapped to the next control inputs $\boldsymbol{\alpha}_{t+1}$.

Following this, the repetitive control process starts. At first, the MP-RNN predicts the $N$ future states of the laser and the corresponding control inputs $\{\boldsymbol{v},\boldsymbol{u}\}_{t_c+1:t_c+N}$, where $t_c$ defines the current time step. These control inputs are optimized to maximize the objective function $O$. To do so, $-\nabla O$ is backpropagated with respect to the control inputs using the BPTT algorithm. The number of optimization steps depends on the previous prediction error. If the prediction is accurate, an appropriate optimization can be expected and, thus, more iteration steps will be performed. However, if the error is increasing, a correct optimization cannot be guaranteed and the number of iteration steps is reduced.
After optimizing, the control inputs $\boldsymbol{u}^*_{t_c+1:t_c+N}$ are used to regulate the MLFL for the next $N$ time steps. If the prediction error exceeds a certain threshold, then the inner loop will be used to stabilize the system.

\section{Results and Performance}

The deep learning and MPC algorithm components outlined in Secs.~3B-D are used to infer the birefringence, build a model of the laser, and produce a control law.  The data generated for building the model is produced by simulating the laser cavity in Sec.~3A.  In our simulations to generate data, the latent variable of birefringence $K$ is varied from $K\in[-0.3,0.3]$.  For each birefringence value, we sweep through the waveplate and polarizer settings in order to determine good regions of mode-locking based upon our objective function $O$.
This data is used for training the VAE and DNN modules for the DL-MPC laser.  All of our code, including the laser simulation engine, deep learning modules and MPC actuation algorithm is provided as open source software in order to promote a reproducible~\cite{baumeister}, and easy to integrate software structure.  Indeed, the python code used for the deep learning module and MPC can be directly integrated with a laser cavity through, for instance, Labview.

We first demonstrate the efficiency of the VAE in identifying the laser birefringence $K$.   
Figure~\ref{fig:VAE_birefringence} demonstrates that the VAE module is able to extract the correct
value of the birefringence $K$ despite its stochastic variability as a function of the number of cavity round trips, or iterations.   Indeed, the birefringence is well tracked, or inferred, by the VAE even as it changes drastically over time.  Note that the VAE produces two latent space outputs:  (i) a single output that tracks the true birefringence (green line), and (ii) a second output capturing what appears to be random, white noise fluctuations.  The output tracking the birefringence can then be used to produce an accurate estimate for the MPC.  Importantly, the result of Fig.~\ref{fig:VAE_birefringence} demonstrates that we can build a good model for the latent variable, i.e. the birefringence $K$.

\begin{figure}[t]
\vspace{-.1in}
\centering
\includegraphics[width=\linewidth]{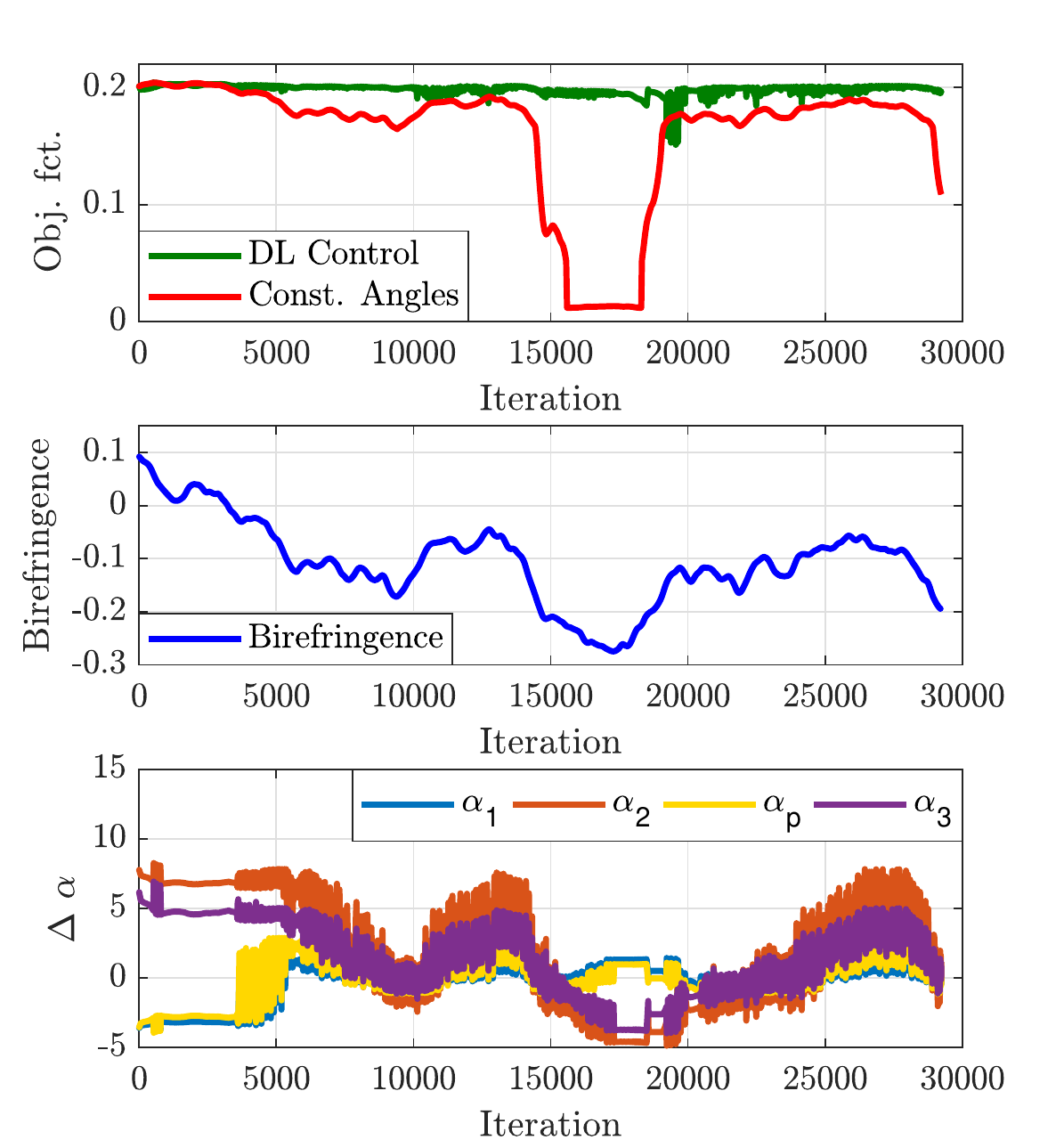}
\vspace{-.25in}
\caption{Same as Fig.~\ref{fig:sinusoidal} but with random changes in birefringence.  The DL-MPC again stabilizes the objective function of the system at a high level.}
\label{fig:random}
\end{figure}

To demonstrate the DL-MPC architecture on the MLFL, we consider two canonical cases.  One in which the birefringence varies smoothly (sinusoidally) in time, and a second in which the birefringence varies stochastically in time.  For both of these scenarios, the modulation of the birefringence in time is equivalent to the birefringence changing as a function of cavity round trips, or iterations.   Figures~\ref{fig:sinusoidal} and \ref{fig:random} compare the performance of the MLFL with and without the DL-MPC, showing that the DL-MPC keeps the MLFL operating at peak performance while a passive MLFL drops out of mode-locking due to the changes in birefringence.  The specific birefringence changes are shown in the middle panels while the evolution of the polarizers and waveplates induced by the DL-MPC are shown in the bottom panel.  These simulated experiments clearly demonstrate the ability of the DL-MPC architecture to learn a model for the unknown birefringence and the control of the waveplates and polarizers.

\section{Conclusion}

Machine learning is revolutionizing science and technology, with deep learning providing the most compelling and successful mathematical architecture available today for model inference.  Indeed, deep learning is a foundational technology for self-driving autonomous vehicles that are already in limited use today.  The mode-locked fiber laser is a mature technology that is broadly used for commercial and scientific purposes.  Yet remarkably, only recently have efforts been made to build self-tuning systems~\cite{Brunton2013jqe,Fu2014oe,Brunton2014ieeejstqe,Andral2015optica,andral2016toward,Woodward2016sr}.  
In this work, we integrate two dominant and leading paradigms for control (MPC) and learning (DNNs) in order to demonstrate a robust and stable method for achieving self-tuning performance in a MLFL.  We additionally provide all the code base and algorithms used here as open source in order to allow researchers to directly build the DL-MPC on their own MLFLs~\cite{baumeister}.  More broadly, the algorithm can be modified to self-tune a broad array of optical systems. 

Although no new optical physics is claimed to have been explored in this paper, the DL-MPC architecture provides a principled method by which latent (unknown or unmeasured) physics can be inferred along with a module for predicting the dynamics of the system.  As more complicated physical and optical systems emerge for technological consideration, the mathematical architecture provided here can provide a robust method for exploring and quantifying new physics.  It can also be used in a {\em grey box} fashion whereby many physical effects are known, and only some additional physical effects must be discovered, parametrized or inferred.  More broadly, it would be truly remarkable if the computer science and machine learning community engenders a fully self-driving car before the optical sciences community learns to self-tune a laser, especially as the input and output space are trivial in comparison to autonomous vehicles.   For many of the physical sciences, it is time to embrace the full potential of machine learning for data-driven discovery of physical principles.

\section*{Funding Information}
SLB acknowledges support from the Army Research Office Young Investigator Program (W911NF-17-1-0422).  
JNK acknowledges support from the Air Force Office of Scientific Research (FA9550-12-1-0253 and FA9550-17-1-0200).

%
%
%
%
%
%
%
\small
\bibliography{sample}

\bibliographyfullrefs{sample}


\normalsize
\section*{Appendix}
\begin{figure}[b]
\centering
\includegraphics[width=\linewidth]{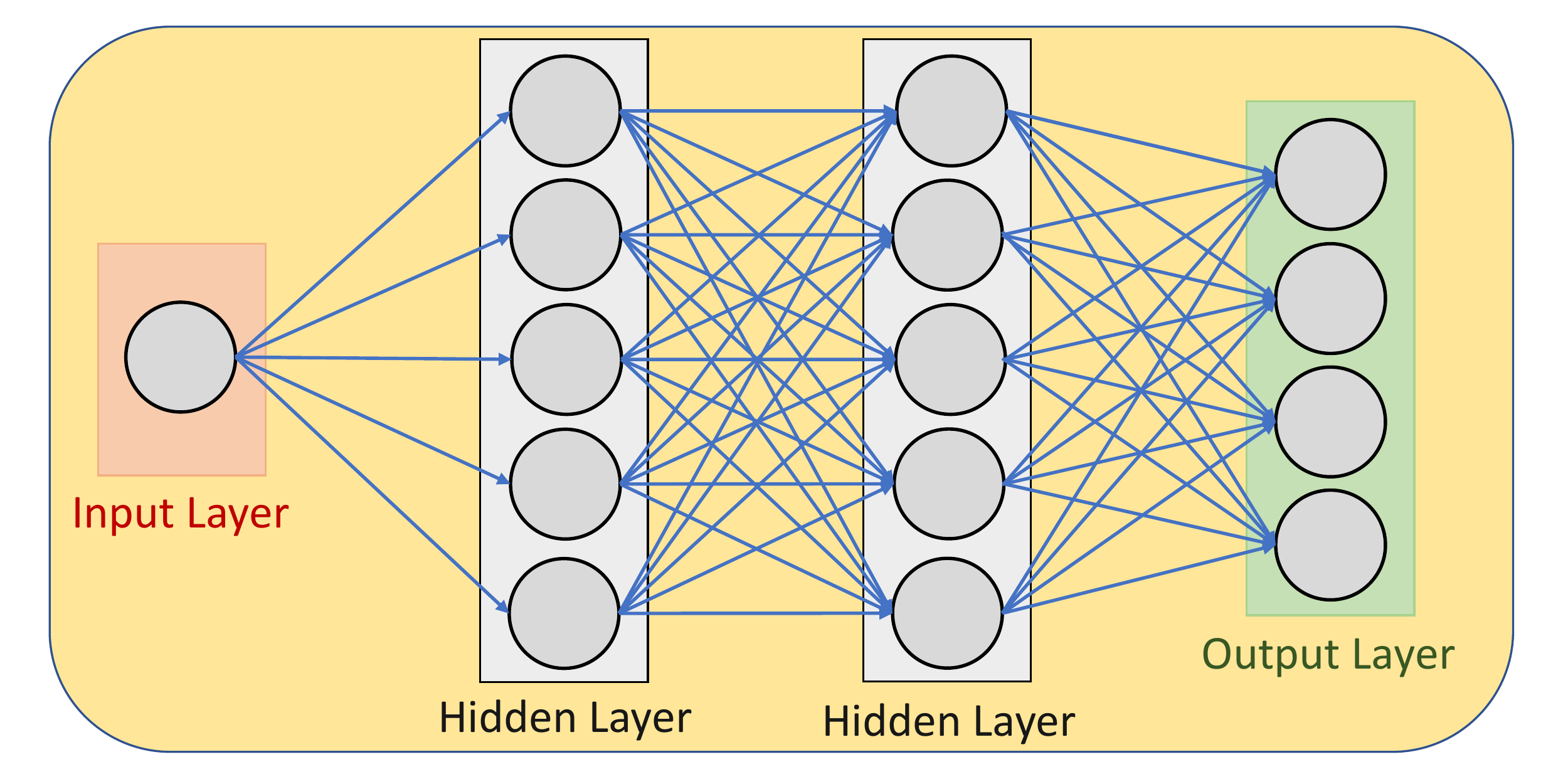}
\caption{A fully connected deep neural network to map the latent variable $K$ to good initial control inputs $\boldsymbol{u}$.}
\label{fig:DNN}
\end{figure}

Figure \ref{fig:DNN} shows that a deep neural network used to map the latent birefringence. 

\paragraph{RNN Cell}

%
The equations used to capture the long-term dynamics are the following:
\begin{align}
&\hspace*{-.1in} \boldsymbol{h}_{l,past} \!=\! \texttt{relu}\!\left(\sum\limits_{i} \boldsymbol{W}^i_{l,past} \boldsymbol{x}^i_{t-2b:t-b-1} + \boldsymbol{b}_{l,past}\right),\\
&\hspace*{-.1in}\boldsymbol{h}_{l,current} \!=\! \texttt{relu}\!\left(\sum\limits_{i} \boldsymbol{W}^i_{l,current} \boldsymbol{x}^i_{t-b:t} + \boldsymbol{b}_{l,current}\right),\\
&\hspace*{-.1in}\boldsymbol{l}_{current} \!=\! \texttt{relu}\!\left(\sum\limits_{i} \boldsymbol{W}^i_{hl} \boldsymbol{h}^i_{l,past} \boldsymbol{h}^i_{l,current} + \sum\limits_{k} \boldsymbol{W}^k_{ll} \boldsymbol{l}^k_{past} + \boldsymbol{b}_{hl}\right),
\end{align}
where the weights are $\boldsymbol{W}_{l,past}$ and $\boldsymbol{W}_{l,current} \in N_x \times N_h$, $\boldsymbol{W}_{hl} \in N_h \times N_l$, $\boldsymbol{W}_{ll} \in N_l \times N_l$ and the biases are $\boldsymbol{b}_{l,past}$  and $\boldsymbol{b}_{l,current} \in N_h$ and $\boldsymbol{b}_{hl} \in N_l$. Take $N_x$ as the number of inputs $\boldsymbol{x}$, $N_h$ as the number of hidden features $\boldsymbol{h}$ and $N_l$ as number of latent features $\boldsymbol{l}$. Since these equations capture the long-term dynamics, its information is forwarded to the next cell, i.e. the output $l_{current,cell_j}$ is forwarded to the next cell where it becomes $l_{past,cell_{j+1}}$.

The second part of the RNN integrates spontaneous changes of the dynamics which are not captured in the long-term dynamics. This is important in case the system is required to respond quickly to unexpected behavior of the system.  So then:
\begin{equation}
h_{current} = \texttt{relu}\left(\sum\limits_{i} \boldsymbol{W}^i_{current} \boldsymbol{x}^i_{t-b:t} + \boldsymbol{b}_{current}\right),
\end{equation}
where $\boldsymbol{W}_{current} \in N_x \times N_h$ and $\boldsymbol{b}_{current} \in N_h$.

The third part captures the influence of the control inputs $\boldsymbol{u}_{t-b+1:t+1}$ on the future states $\boldsymbol{v}_{t+1}$:
\begin{equation}
h_{future} = \texttt{relu}\left(\sum\limits_{i} \boldsymbol{W}^i_{future} \boldsymbol{u}_{t-b+1:t+1} + \boldsymbol{b}_{future}\right),
\end{equation}
where the weights are $\boldsymbol{W}_{future} \in N_u \times N_h$, the biases are $\boldsymbol{b}_{future} \in N_h$ and $N_u$ is the number of control inputs $\boldsymbol{u}$. Since the future states $\boldsymbol{v}_{t+1}$ strongly depend on $\boldsymbol{u}_{t+1}$, we first predict the future latent variable $K_{t+1}$, which depends on known values:
\begin{align}
\boldsymbol{h}_{latent} &= \texttt{relu}\left(\sum\limits_{i} \boldsymbol{W}^i_l \boldsymbol{l}^i_{current} + \boldsymbol{b}_{latent}\right),\\
\boldsymbol{K}_{t+1} &= \sum\limits_{i} \boldsymbol{W}^i_{Ko} \boldsymbol{h}^i_{latent} + \boldsymbol{b}_{Ko}.
\end{align}
Then, $\boldsymbol{K}_{t+1}$ is mapped to $\boldsymbol{u}_{t+1}$ using $K-{\bf u}$ mapping. Now the computed  $\boldsymbol{u}_{t+1}$ can be used to predict the future states  $\boldsymbol{v}_{t+1}$:

\begin{equation}
\boldsymbol{v}_{t+1} = \sum\limits_{i} \boldsymbol{W}^i_o \boldsymbol{h}^i_{latent} \boldsymbol{h}^i_{current} \boldsymbol{h}^i_{future} + \boldsymbol{b}_o.
\end{equation}
Here, the weights are $\boldsymbol{W}_l \in N_l \times N_h$, $\boldsymbol{W}_{Ko} \in N_h \times N_K$, and $\boldsymbol{W}_o \in N_h \times N_v$ and the biases are $\boldsymbol{b}_{latent} \in N_l$, $\boldsymbol{b}_{Ko} \in N_K$ and $\boldsymbol{b}_o \in N_v$. Take $N_K$ as number of latent variables for $K$ and $N_v$ as the number of system states $\boldsymbol{v}$. For each hidden and latent layer, rectified linear units are chosen as activation functions to restrain the vanishing gradient and the output layer is linear.

\end{document}